\documentclass{article} % For LaTeX2e
\usepackage{iclr2015,times}
\usepackage{hyperref}
\usepackage{url}
\usepackage{hyperref}
\usepackage{url}
\usepackage{times}
\usepackage{calc}
\usepackage{graphicx}
\usepackage{amsmath}
\usepackage{amssymb}
\usepackage{relsize}
\usepackage{multirow}
\usepackage{bm}
\usepackage{rotating}
\usepackage{graphicx}
\usepackage{multicol}
\usepackage{comment}
\usepackage{pgfbaselayers} 
\usepackage{enumitem}
\usepackage[none]{hyphenat}
\usepackage{times}
\usepackage{epsfig}
\usepackage{graphicx}
\usepackage{amsmath}
\usepackage{amssymb}
\usepackage{subfigure}

\title{Compressing Deep Convolutional Networks using Vector Quantization}

\author{Yunchao Gong,~~~Liu Liu \thanks{Currently with Snapchat.},~~~Ming Yang,~~~Lubomir Bourdev\\
Facebook AI Research\\
{\tt\small \{ycgong,lliu,mingyang,lubomir\}@fb.com}
}

\iclrconference % Uncomment if submitted as conference paper instead of workshop

\begin{document}

\maketitle

\begin{abstract}
Deep convolutional neural networks (CNN) has become the most promising method for object recognition, repeatedly demonstrating record breaking results for image classification and object detection in recent years. However, a very deep CNN generally involves many layers with millions of parameters, making the storage of the network model to be extremely large. This prohibits the usage of deep CNNs on resource limited hardware, especially cell phones or other embedded devices. In this paper, we tackle this model storage issue by investigating information theoretical vector quantization methods for compressing the parameters of CNNs. In particular, we have found in terms of compressing the most storage demanding dense connected layers, vector quantization methods have a clear gain over existing matrix factorization methods. Simply applying k-means clustering to the weights or conducting product quantization can lead to a very good balance between model size and recognition accuracy. For the 1000-category classification task in the ImageNet challenge, we are able to achieve 16-24 times compression of the network with only 1\% loss of classification accuracy using the state-of-the-art CNN.
\end{abstract}

\section{Introduction}

Deep convolutional neural networks \citep{Krizhevsky12,lecun90,DBLP:journals/corr/SzegedyLJSRAEVR14,DBLP:journals/corr/SimonyanZ14a} has recently achieved significant progress and have become the gold standard for object recognition, image classification, and retrieval. Almost all of the recent successful recognition systems \citep{Jia13caffe,jia13,simonyan2013deep,sermanet2013overfeat,zeiler2013visualizing,gong14} are built on top of this architecture.  Importing CNN onto embedded platforms \citep{Gokhale13,Culurciello13}, the recent trend toward mobile computing, has a wide range of application impacts. It is especially useful when the bandwidth is limited or photos are not allowed to be sent to servers. However, the size of the CNN models are typically very large (e.g. more than 200M bytes), which limits the applicability of such models on the embedded platform. For example, it will be almost impossible for users to download an iPhone application with more than 20M. Thus, in order to apply neural network methods to embedded platforms, one important research problem is how to compress parameters to reduce storage requirements.

In this article we mainly consider compressing CNNs for computer vision tasks. For example, a typical CNN \citep{Jia13caffe,sermanet2013overfeat,zeiler2013visualizing} that works well for object recognition contains eight layers (five convolutional layers and three dense connected layers) and a huge number of parameters (e.g. $10^8$) in order to produce state-of-the-art results. Because as is widely known, the parameters are heavily over-parameterized \citep{NIPS2013_5025}, it is very interesting to investigate whether we can compress these parameters by exploring their structure. Here, we are mainly interested in compressing the parameters to reduce storage instead of speeding up the testing time \citep{Denton}. For a typical network described in \citep{zeiler2013visualizing}, about 90\% of the storage is taken up by the dense connected layers; more than 90\% of the running time is taken by the convolutional layers. Therefore, we shall focus upon how to compress the dense connected layers to reduce storage of neural networks.

%\begin{figure}[t]
%\centering
%\subfigure[Storage.]{
 % \includegraphics[width= 1.55in,trim = 5mm 65mm 5mm 75mm]{figure/storage.pdf}}
%\subfigure[Testing Time.]{
 % \includegraphics[width= 1.55in,trim = 5mm 65mm 5mm 75mm]{figure/time.pdf}}  
  %\caption{Storage and test time spent on convolutional layers and dense connected layers.}\label{stats}
%\end{figure}

A few early works on compressing CNNs have been published; however, their focus is different from ours. The most closely related one, \cite{Denton}, explored matrix factorization methods for speeding up CNN testing time. These researchers showed that by exploring the linear structure of CNN parameters (in particular, convolutional layers), CNN testing time can be sped up by as much as 200\% while keeping the accuracy within 1\% of the original model. Another similar work on speeding up CNN is \citep{Jaderberg14}, in which the authors described several reconstruction methods for approximating the filers in convolutional layers. Their goal of both works is complimentary to ours, in that they focus on compressing convolutional layers for speeding up CNN. Our focus, however, is on compressing dense connected layers in order to reduce the size of the model.  

In this work, instead of the traditional matrix factorization methods considered in \citep{Denton,Jaderberg14}, we mainly consider a series of information theoretical vector quantization methods \citep{jegou2010b,rq} for compressing dense connected layers. For example, we consider binarizing the parameters, scalar quantization using $k$means, and structured quantization using product quantization or residual quantization. Surprisingly, we have found that simply applying $k$means-based scalar quantization achieves very impressive results, and that this approach is better than matrix factorization methods. Structured quantization methods can give additional gain by exploring redundancy in the parameters. To our knowledge, this is the first work to systematically study different vector quantization methods for compressing CNN parameters.

This paper makes the following contributions: 1) We are among the first to systematically explore vector quantization methods for compressing the dense connected layers of deep CNNs to reduce storage; 2) We have performed a comprehensive evaluation of different vector quantization methods, which has shown in particular that structured quantization such as product quantization works significantly better than other methods; 3) We have performed experiments on other tasks such as image retrieval, to verify the generalization ability of the compressed model.
  
\section{Related Work\label{related}}

Deep convolutional neural network has achieved great successes in image classification \citep{Krizhevsky12,Jia13caffe,jia13,simonyan2013deep,sermanet2013overfeat,zeiler2013visualizing}, object detection \citep{girshick14CVPR}, and image retrieval \citep{DBLP:journals/corr/RazavianASC14,gong14}. With the great progress in this area, the state-of-the-art image classifier can achieve 94\% top five accuracy on the ILSVRC2014 dataset with 1000 object classes,  and is already very close human performance. Such great success ignites interest in adopting CNNs to real world applications. For example, CNN has already been applied to object classification, scene classification, and indoor scene classification. It has also been applied to image retrieval, with impressive results. 
 
As discussed in the above section, a state-of-the-art CNN usually involves hundreds of millions of parameters, which require huge storage for the model that is difficult to achieve. The bottleneck comes from model storage and testing speed. Several works have been published on speeding up CNN prediction speed. \cite{Vanhoucke}, who explored the properties of CPU to speed up the execution of CNN, particularly focused on the aligning of memory and SIMD operations to boost matrix operations. \cite{Mathieu} showed that the convolutional operation can be efficiently carried out in the Fourier domain, which leads to a speed-up of 200\%. Two very recent works by \cite{Denton} and  \cite{Jaderberg14}, which explored the use of linear matrix factorization methods for speeding up convolutional layers, showed a 200\% speed-up with little compromise of classification performance. Almost all of the above mentioned works focus on making the prediction speed of CNN faster; little work has been specifically devoted to making CNN models smaller.

The use of vector quantization methods to compress CNN parameters is mainly inspired by the work of \cite{NIPS2013_5025}, who demonstrate the redundancies in neural network parameters. They show that the weights within one layer can be accurately predicted by a small (~5\%) subset of the parameters, which indicates that the neural network is over-parameterized. These results motivate us to apply vector quantization methods to explore the redundancy in parameter space. In particular, our paper can be viewed as a compression realization of the parameter prediction results reported in \cite{NIPS2013_5025}. Somewhat surprisingly, we have found very similar results to those of \cite{NIPS2013_5025}, in that we are able to compress the parameters about 20 times with little decrease of performance. This result further confirms the interesting empirical findings in \cite{NIPS2013_5025}.

\section{Compress Dense Connected Layers}

In this section, we consider two classes of methods for compressing the parameters in dense connected layers. We first consider the matrix factorization methods, and then introduce the vector quantization methods.

\subsection{Matrix Factorization Methods}

We first consider matrix factorization methods, which have been widely used to speed up CNN \citep{Denton} as well as for compressing parameters in linear models \citep{Denton}. In particular, we consider using singular-value decomposition (SVD) to factorize the parameter matrix. Given the parameter $W \in R^{m \times n}$ in one dense connected layer, we factorize it as 
\begin{equation}
W = USV^T,
\end{equation}
where $U \in R^{m\times m}$ and $V \in R^{n\times n}$ are two dense orthogonal matrices and $S \in R^{m \times n}$ is a diagonal matrix. In order to approximate $W$ using two much smaller matrices, we can pick the top $k$ singular vectors in $U$ and $V$ with corresponding eigenvalue in $S$, to reconstruct $W$:
\begin{equation}
\hat{W} = \hat{U}\hat{S}\hat{V}^T,
\end{equation}
where $\hat{U} \in R^{m \times k}$ and $\hat{V} \in R^{n \times k}$ are two submatrices that correspond to the leading $k$ singular vectors in $U$ and $V$. The diagonal elements in $\hat{S} \in R^{k \times k}$ correspond to the largest $k$ singular values. The approximation of SVD is controlled by the decay along the eigenvalues in $S$. The SVD method is optimal in the sense of a Frobenius norm, which minimizes the MSE error between the approximated matrix $\hat{W}$ and original $W$.
The two low-rank matrices $\hat{U}$ and $\hat{V}$, as well as the eigenvalues, must be stored. So the compression rate given $m$, $n$, and $k$ is computed as ${m n}/{k (m+n+1)}$.

\subsection{Vector Quantization Methods}

\subsubsection{Binarization}

We start with the simplest way to quantize the parameter matrices. Given the parameter $W$, we take the sign of the matrix:
\begin{equation}
\hat{W}_{ij} = \left\{ \begin{array}{ll}
         1 & \mbox{if $W_{ij} \geq 0$},\\
        -1 & \mbox{if $W_{ij} < 0$}.\end{array} \right.
\end{equation}
This method is mainly inspired by Dropconnect \citep{wan13}, which randomly sets part of the parameters (neurons) to 0 during training. Here, we are taking a more aggressive approach by turning each neuron on if it is positive and turning it off when it is negative. From a geometric point of view, assuming that dense connected layers are a set of hyperplanes, we are actually rounding each hyperplane to its nearest coordinate. This method will compress the data by 32 times, since each neuron is represented by one bit.

\subsubsection{Scalar Quantization using $k$means}

Another simple method is to perform scalar quantization to the parameters. For $W \in R^{m\times n}$, we can collect all its scalar values as $\boldsymbol w \in R^{1 \times mn}$, and perform $k$means clustering to the values:
\begin{equation}
\min \sum_i^{mn} \sum_{j}^{k}  \|w_i - c_j\|_2^2,
\end{equation}
where $w$ and $c$ are both scalars. After the clustering, each value in $\boldsymbol w$ is assigned a cluster index, and a codebook can be formed of $\boldsymbol c^{1\times k}$ the cluster centers. During the prediction, we can directly look up the values for each $w_{ij}$ in $\boldsymbol c$. Thus, the reconstructed matrix is:
\begin{equation}
\hat{W}_{ij} = c_{z}, ~~\operatorname{where}~~\min_{z} \|W_{ij} - c_{z}\|_2^2.
\end{equation}

For this approach, we only need to store the indexes and the codebook as the parameters. Given the $k$ centers, we only need $\log_2(k)$ bits to encode the centers. For example, if we use $k=256$ centers, only need 8 bits are needed per cluster index. Thus, the compression rate is $32 / \log_2(k)$, assuming that we use floating numbers for the original $W$ and assuming that the codebook itself is negligible. Despite the simplicity of this approach, our experiments showed that this approach gives surprisingly good performance for compressing parameters.

\subsubsection{Product Quantization\label{pq}}

Next, we consider structured vector quantization methods for compressing the parameters. In particular, we consider the use of product quantization (PQ) \citep{jegou2010b}, which explores the redundancy of structures in vector space. The basic idea of PQ is to partition the 
vector space into many disjoint subspaces, and perform quantization in each subspace. Due to its assumption that the vectors in each subspace are heavily redundant and by performing quantization in each subspace, we are able to better explore the redundancy structure. Specifically, given the matrix $W$, we partition it colum-wise into several submatrices:
\begin{equation}
W = [W^1, W^2, \ldots, W^s],
\end{equation}
where $W^i \in R^{m\times (n/s)}$ assuming $n$ is divisible by $s$. We can perform a $k$means clustering to each submatrix $W^i$:
\begin{equation}
\min \sum_z^{m} \sum_{j}^{k}  \|\boldsymbol w^{i}_{z} - \boldsymbol c_j^i\|_2^2,
\end{equation}
where $\boldsymbol w^{i}_{z}$ denotes the $z$th row of sub-matrix $W^i$, and $\boldsymbol c_j^i$ denotes the $j$th row of sub-codebook $C^i \in R^{k\times (n/s)}$. For each sub-vector $\boldsymbol w^{i}_{z} $, we need only to store its corresponding cluster index and the codebooks. Thus, the reconstructed matrix is:
\begin{equation}
\hat{W} = [\hat{W}^1, \hat{W}^2, \ldots, \hat{W}^s],~~~ \operatorname{where}
\end{equation}
\begin{equation}
\hat{\boldsymbol w}_{j}^i = \boldsymbol c_{j}^i, ~~\operatorname{where}~~\min_{j} \|\boldsymbol w_{z}^i  - \boldsymbol c^i_{j}\|_2^2.
\end{equation}
Note that PQ can be applied to either the x-axis or the y-axis of the matrix. In our experiments, we evaluated both cases. 

For this approach, we need to store the cluster indexes and codebooks for each subvector. In particular, in contrast to the scalar quantization case, the codebook here is not negligible. The compression rate for this method is $(32mn) / (32kn + \log_2(k)ms)$.

\subsubsection{Residual Quantization}

The third quantization method we consider is residual quantization \citep{rq}, which is another form of structured quantization. The basic idea is to first quantize the vectors into $k$ centers and then to recursively quantize the residuals. For example, given a set of vectors $\boldsymbol w_i, i\in 1,\ldots,m$ at the first stage, we begin by quantizing them into $k$ different vectors using $k$means clustering:
\begin{equation}
\min \sum_z^{m} \sum_{j}^{k}  \|\boldsymbol w_{z} - \boldsymbol c^{1}_j\|_2^2,
\end{equation}
Every vector $\boldsymbol w_z$ will be represented by its closest center $\boldsymbol c_j^1$. Next, we compute the residual $\boldsymbol r_{z}^1$ between $\boldsymbol w_z$ and $\boldsymbol c_j^1$ for all the data points and recursively quantize the residual vectors $\boldsymbol r_{z}^1$ into $k$ different code words $\boldsymbol c_j^2$. Finally, a vector can be reconstructed by adding its corresponding centers at each stage:
\begin{equation}
\hat{\boldsymbol w_z} = \boldsymbol c_j^1 + \boldsymbol c_j^2 + \ldots, \boldsymbol c_j^t,
\end{equation}
given we have recursively performed $t$ iterations.

We need to store all the codebooks for each iteration, which potentially needs large a amount of memory. The compression rate is $m/(tk + \log_2(k)tn)$.

\subsubsection{Other Methods and Discussion}

The abovementioned (KM, PQ, and RQ) are three different kinds of vector quantization methods for compressing matrices. KM only captures the redundancy for each neuron (single scalar); PQ explores some local redundancy structure; and RQ tries to explore the global redundancy structure between weight vectors. It will be interesting to investigate what kinds of redundancies are present in the behavior of the learned parameters. 

Many learning-based binarization or  product quantization methods are available, such as Spectral Hashing \citep{weiss08}, Iterative Quantization \citep{gongpami}, and Catesian $k$means \citep{norouzi13}, among others. However, they are not suitable for this particular task because we need to store the learned parameter matrix (e.g., a rotation matrix), which is very large. For this reason, we do not consider the other methods in this paper. It is also interesting to explore the structure in $W$ when performing PQ. In particular, because the $W$ is learned on a set of outputs of different filters, grouping together output from specific filters or grouping together specific dimensions from different filters might be  interesting. However, in our preliminary investigation, we did not find any improvement by exploring such structure when doing PQ. Therefore, we grouped these dimensions by default order.

\section{Experiments}

\subsection{Experimental Setting}

We evaluated these different methods on the ILSVRC2012 benchmark image classification dataset. This dataset contains more than 1 million training images from 1000 object categories. It also has a validation set of 20,000 images, in categories that contain 20 images each. We trained on the standard training set, performed the compression to the parameters, and tested on the validation set.

The convolutional neural network we used, from \cite{zeiler2013visualizing}, contains 5 convolutional layers and 3 dense connected layers. All of the input images were first resized to minimal dimensions of 257, after which we performed random cropping to 225$\times225$ patches. Then the images were fed into 5 different convolutional layers with respective filter sizes of 7, 5, 3, 3, and 3. The first two convolutional layers were followed by a local response normalization layer and a max pooling layer. At this point we have obtained three fully connected layers of sizes $9216\times2048$, $2048\times2048$ and $2048\times1000$. The nonlinear function we used here was RELU. The network was trained on 1 GPU for about 5 days after 70 epochs. The learning rate started at 0.02 and halved every 5-10 epochs; the weight decay was set to 0.0005; and momentum was set to 0.9.

To evaluate the different methods, we used the classification accuracy on the validation set as the evaluation protocol. We used both the accuracy@1 and accuracy@5 to evaluate different parameter compression methods. The goal was either to achieve higher compression rate with same accuracy or higher accuracy with same compression rate.

\subsection{Analysis of Product Quantization}

We first performed an analysis on the PQ-based compression of the parameters, because PQ has several different parameters. We used a different number of clusters $k=4,8,16$ (corresponds to $2,3,4$ bits) for each segment. For each fixed $k$, we show results for different segment dimension (column) sizes $s = 1,2,3,4$, which changed the compression rate from lower to higher.  As mentioned in section \ref{pq}, we were able to perform PQ for either the x-axis or the y-axis and therefore show results for both cases. We shall compare different PQ methods when we align the segment size, and also compare them with the aligned compression rate.

\begin{figure*}[!t]
\centering
\subfigure[PQ for X-axis]{
  \includegraphics[width= 2.5in,trim = 20mm 60mm 0mm 80mm]{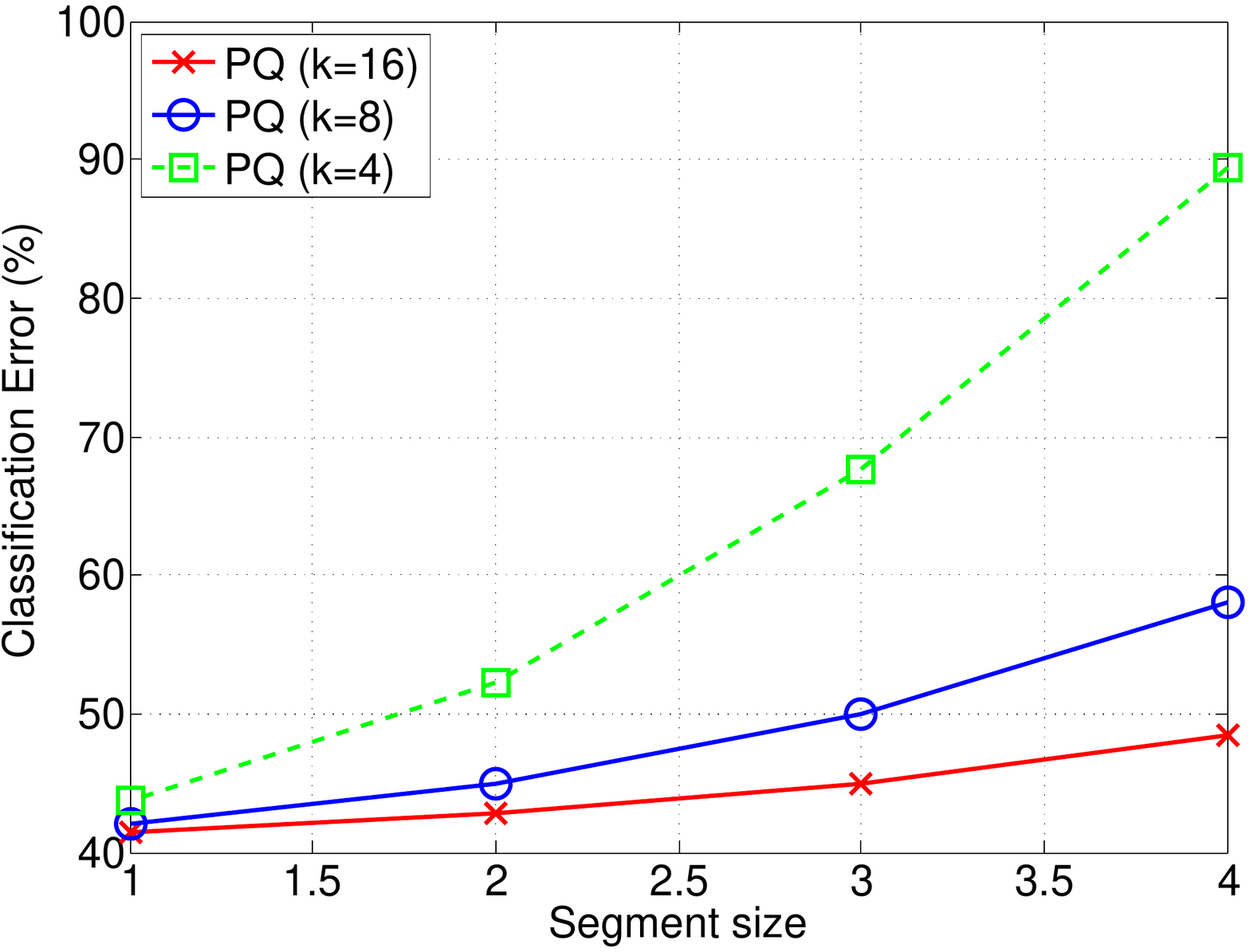}}
\subfigure[PQ for Y-axis]{
  \includegraphics[width= 2.5in,trim = 0mm 60mm 20mm 80mm]{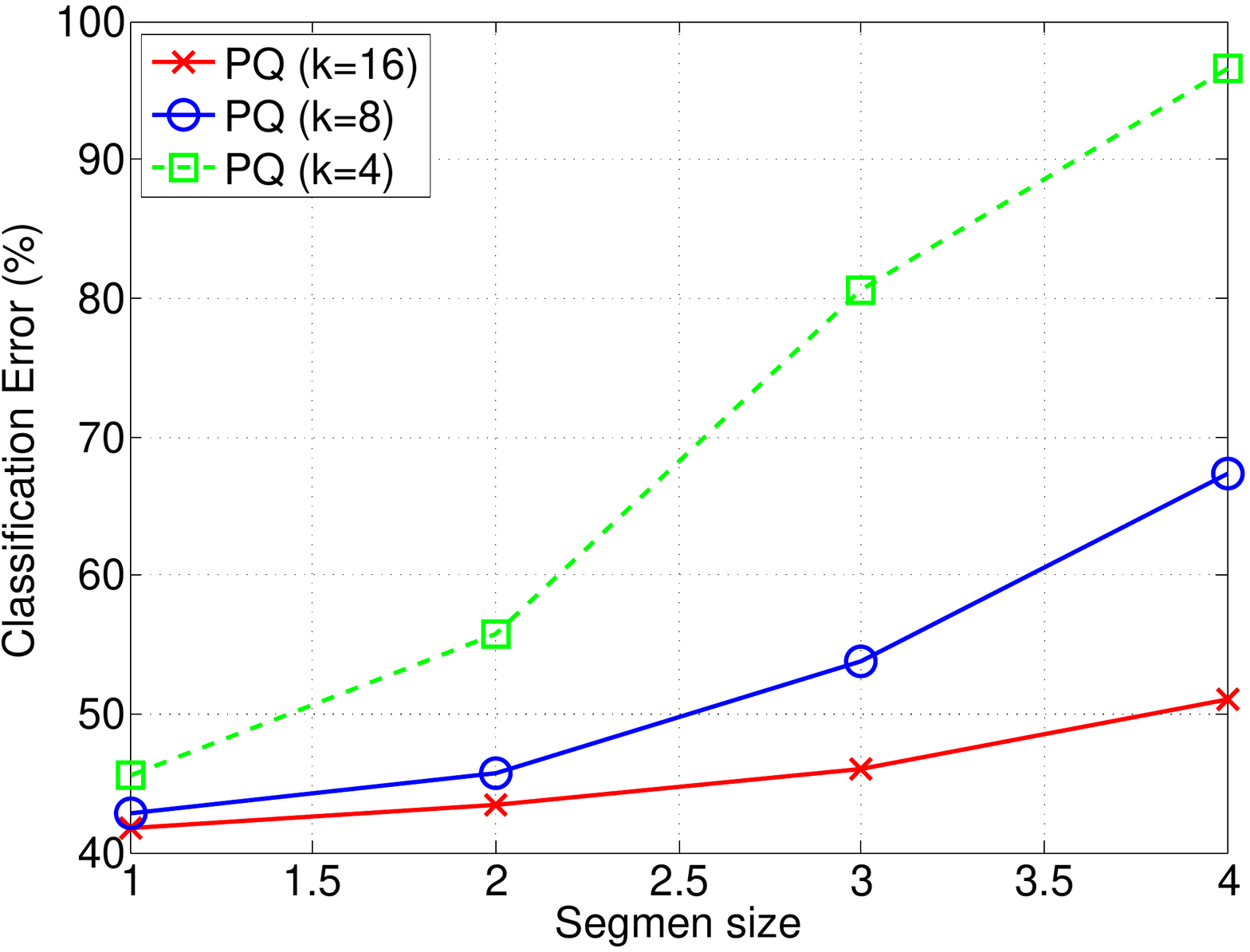}}  \vspace{-3mm}
  \caption{Comparison of PQ compression with aligned segment size for accuracy@1.}\label{pq1}
\end{figure*}

\begin{figure*}[!t]
\centering
\subfigure[PQ for X-axis]{
  \includegraphics[width= 2.5in,trim = 20mm 60mm 0mm 80mm]{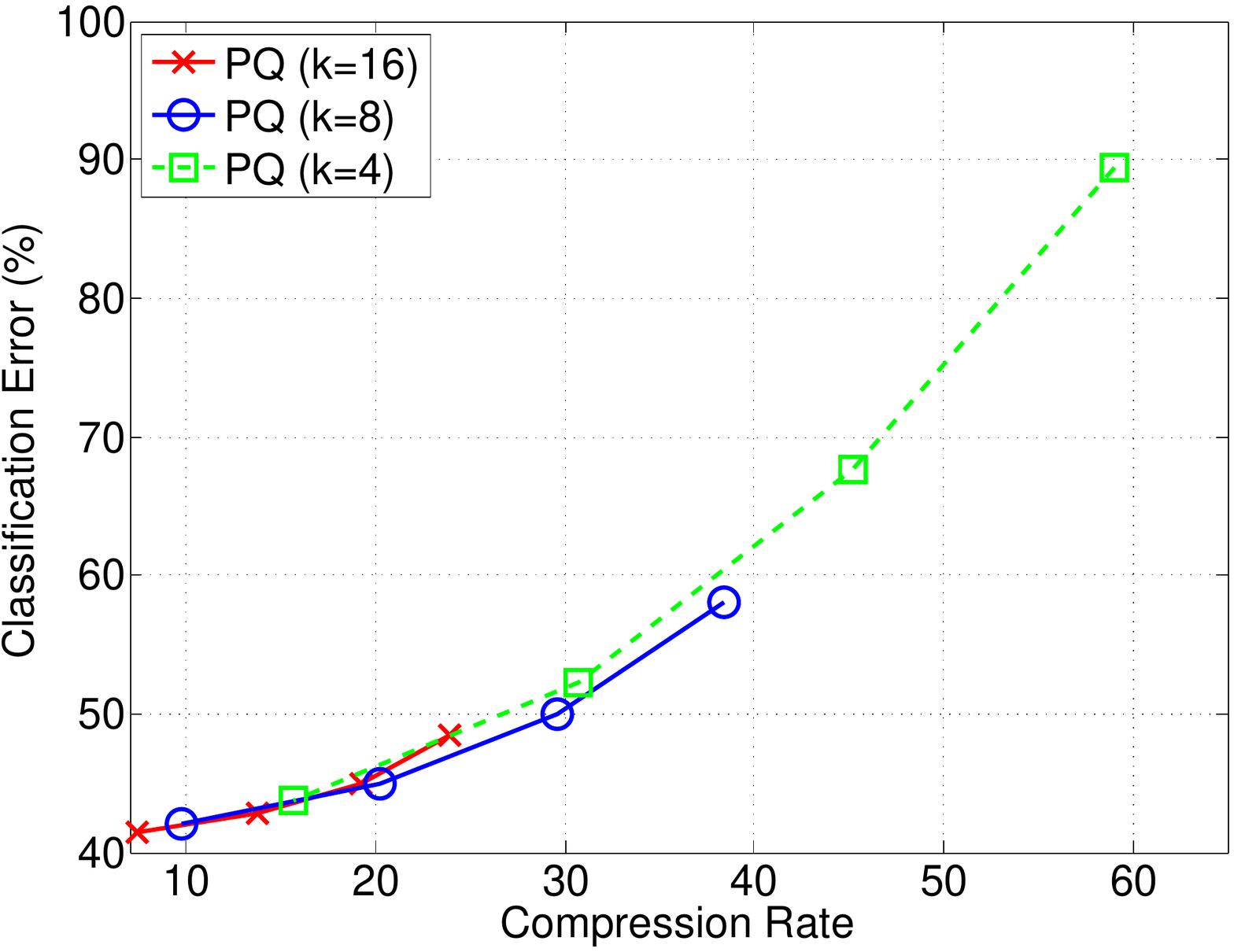}}
\subfigure[PQ for Y-axis]{
  \includegraphics[width= 2.5in,trim = 0mm 60mm 20mm 80mm]{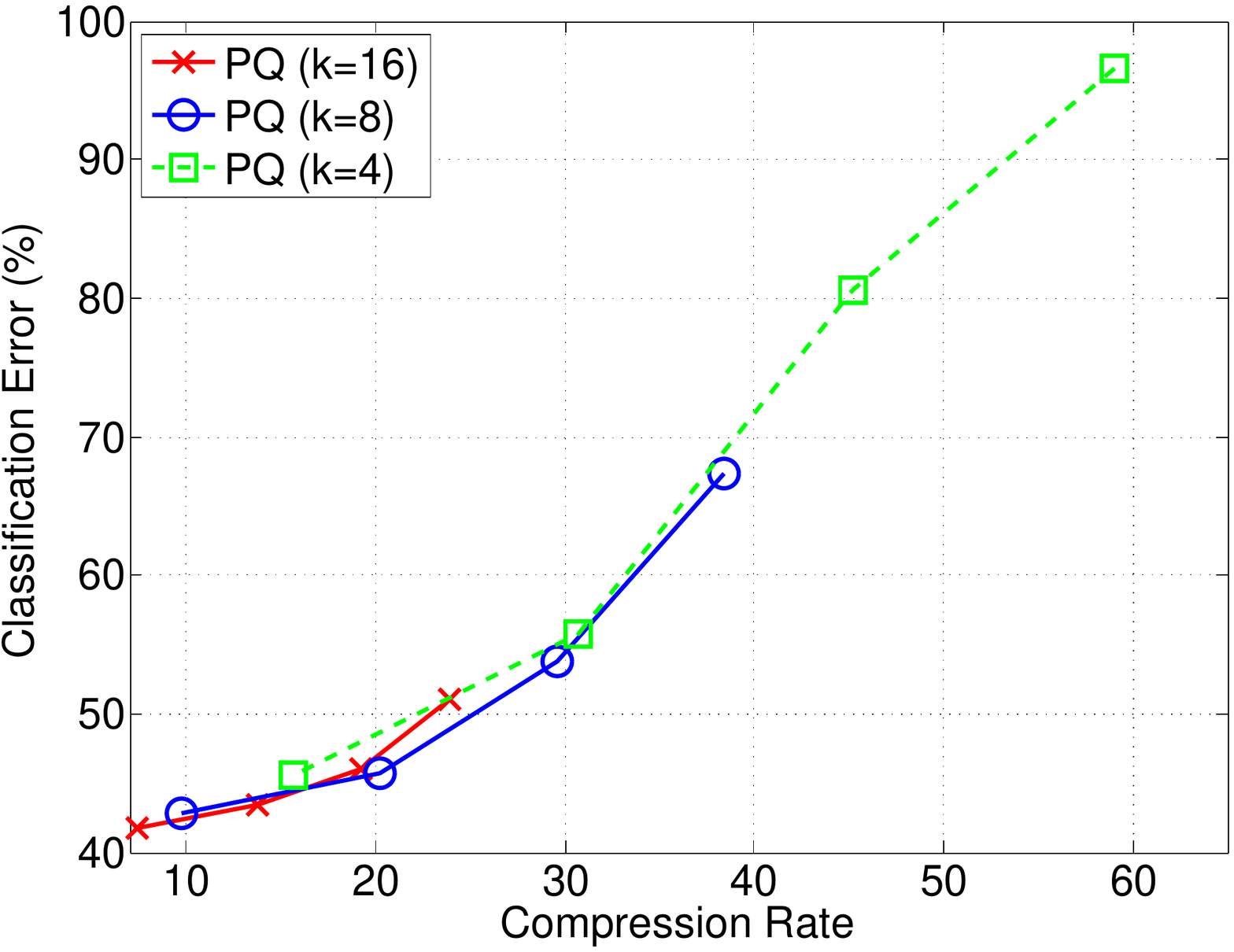}}   \vspace{-3mm}
  \caption{Comparison of PQ compression with aligned compression rate for accuracy@1. We can clearly find when taking codebook size into account, using more centers do not necessarily lead to better accuracy with same compression rate. See text for detailed discussion.}\label{pq2}
\end{figure*}

The results for accuracy@1 are reported in Figure \ref{pq1} and Figure \ref{pq2} for different axis alignments.  From the results in Figure \ref{pq1}, we see that by using more centers $k$ and smaller segment size $s$, we were able to obtain smaller classification error. For example, in Figure \ref{pq1}, the red curve always has much smaller classification error than other methods. This result is consistent with previous observations for PQ (i.e., that using more centers in each segment can usually obtain a lower rate of quantization error). However, when we took the size of the codebook into account and measured the accuracy for the same compression rate, as in Figure \ref{pq2}, we found using more centers is not always helpful because they will aggressively increase the codebook size. For example, when we used $k=16$ centers, the classification error was clearly not lower than when we used fewer number of clusters (e.g. $k=8$). This difference is because the codebook itself is using too much storage space and and makes the compression rate very low.  When we compare the results of compressing the x-axis and y-axis, the results for the x-axis are slightly better. This improvement is probably because there are more dimensions for the x-axis, which makes the codebook size larger and thereby reduces the loss of information. In the next set of experiments, we fixed the number of centers to 8 (3 bits per segment) because this approach achieves a good balance between compression rate and accuracy.

\subsection{Overall Comparison}

Section 3 below contains a comparison of all of the quantization methods we introduce herein. Similarly to the above sections, we present the classification errors with respect to the compression rate. For binary quantization, which has no parameter to tune, the compression rate was 32. For PQ, we used 8 centers per segment and varied the dimension (from 1 to 4) of each segment to achieve different compression rates. For $k$means (KM), we varied the number of clusters from 1 to 32 to achieve a compression rate between 32 and 1. For SVD, we vary the output dimensionality to achieve different compression rates. The performance of RQ was unsatisfactory; here, we report results only for 256 centers with $t=2,3$ iterations. We also experimented with lower numbers of centers for RQ, but found the performance to  be even worse.

\begin{figure*}[t]
\centering
\subfigure[Accuracy@1]{
  \includegraphics[width= 2.5in,trim = 20mm 60mm 0mm 85mm]{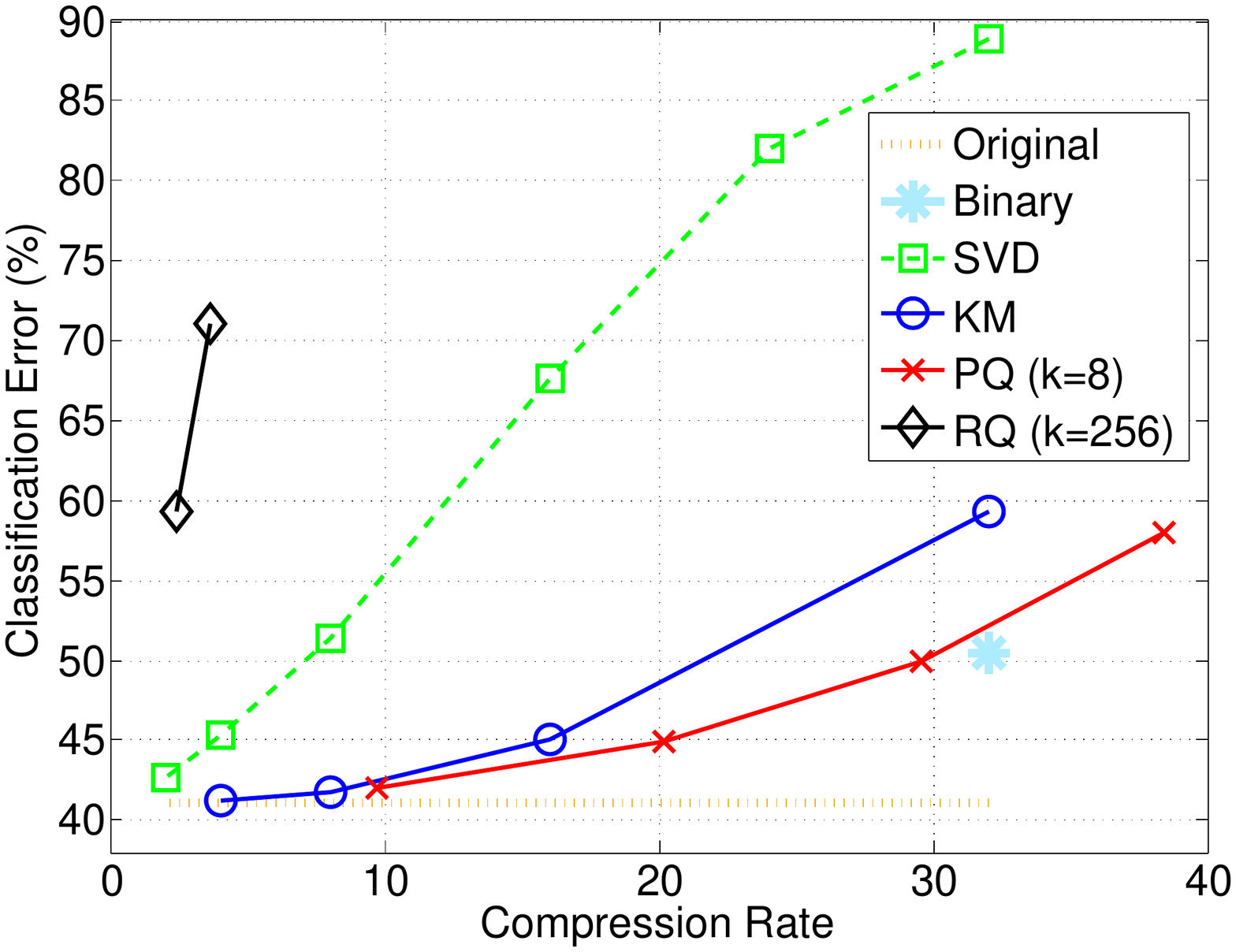}}
\subfigure[Accuracy@5]{
  \includegraphics[width= 2.5in,trim = 0mm 60mm 20mm 85mm]{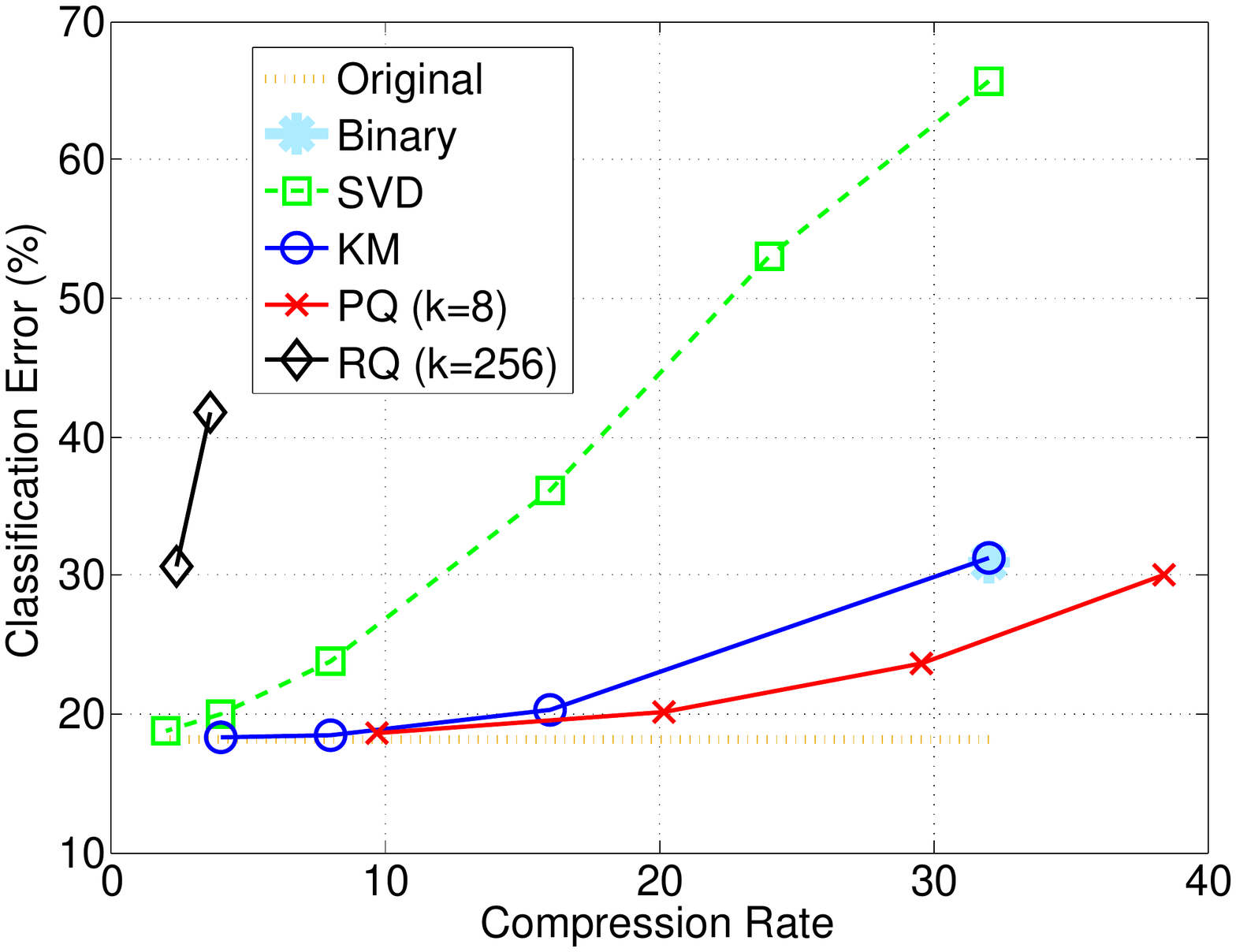}}  \vspace{-3mm}
  \caption{Comparison of different compression methods on ILSVRC dataset.}\label{all}
\end{figure*}

Both accuracy@1 and accuracy@5 are shown in Figure \ref{all}, where we see that  the trend is consistent for both figures. In particular, SVD  achieves impressive results for compressing and speeding up convolutional layers \citep{Denton} but does not work well for compressing dense connected layers. This difference is mainly because the two factorized matrices still need to be stored for SVD, which is not optimized for saving storage.  Somewhat surprisingly, $k$means, despite its simplicity, works well for this task. We were able to achieve a 4-8 times compression rate with KM while keeping the accuracy loss within 1\%. Applying structured quantization methods such as PQ can further improve performance beyond our results for KM.  For RQ, we found it was not able to achieve a very high compression rate, mainly because the codebook size was too big. Given the same compression rate, its accuracy was also much worse than the other methods. Finally, the simplest binarization method worked reasonably well. We were not able to vary its compression rate, but given the same compression rate its performance was comparable to KM or PQ. In addition, its storage is quite simple. Therefore, this method is also a good choice when the goal is to compress data very aggressively. 
 
The comparison among KM, PQ, and RQ suggests some interesting insights. First, KM works reasonably well and can achieve a descent compression rate without sacrificing performance. These results suggests that there is considerable redundancy between each single neuron. Applying PQ works even better, which means that there are very meaningful sub-vector local structures in these weight matrices.  RQ works extremely poorly for such a task, which probably means there are few global structures in these weight vectors. 

\subsection{Analysis of Single Layer Error}

\begin{figure*}[t]
\centering
\subfigure[SVD]{
  \includegraphics[width= 1.7in,trim = 10mm 62mm 10mm 75mm]{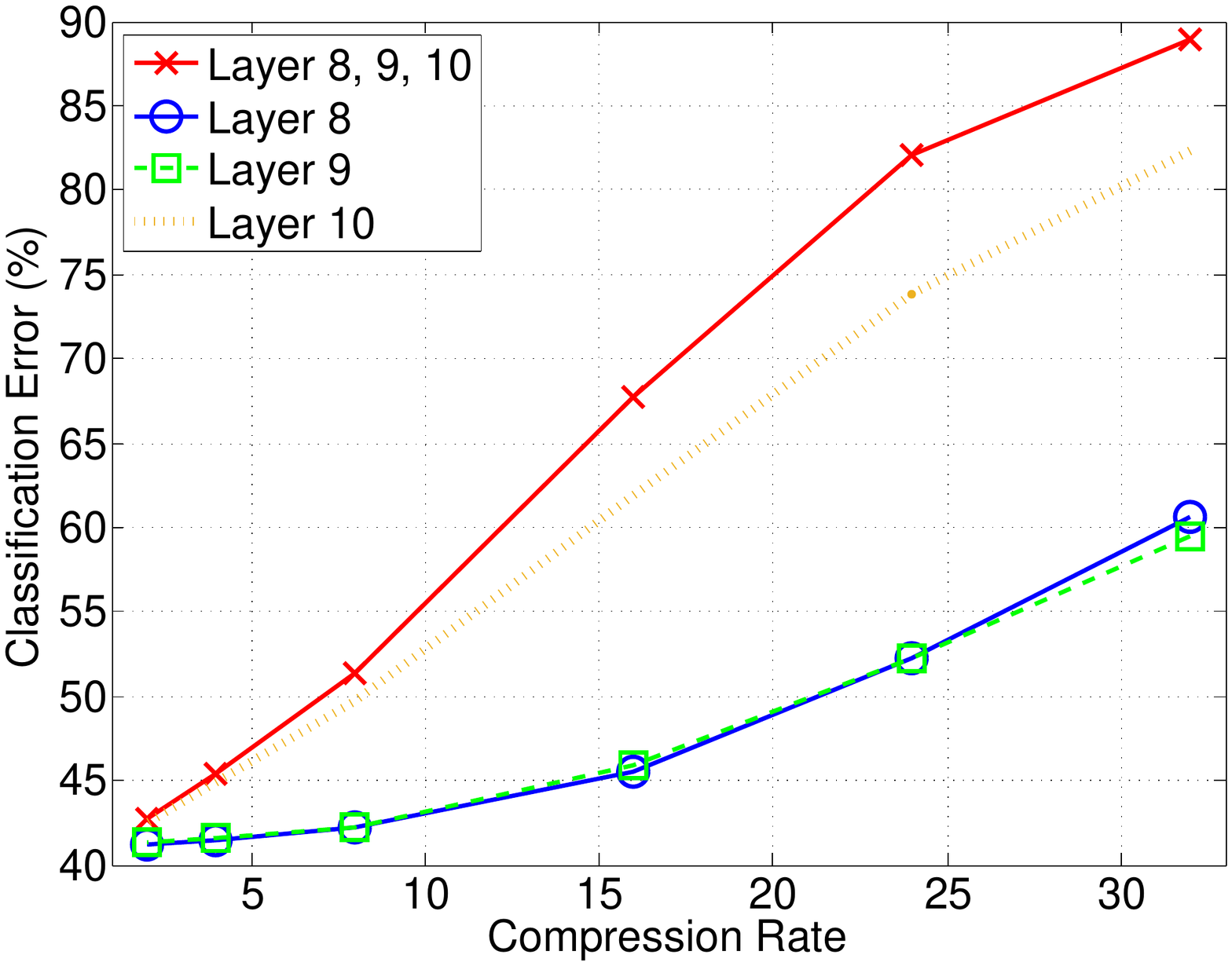}}    
\subfigure[KM]{
  \includegraphics[width= 1.7in,trim = 10mm 62mm 10mm 75mm]{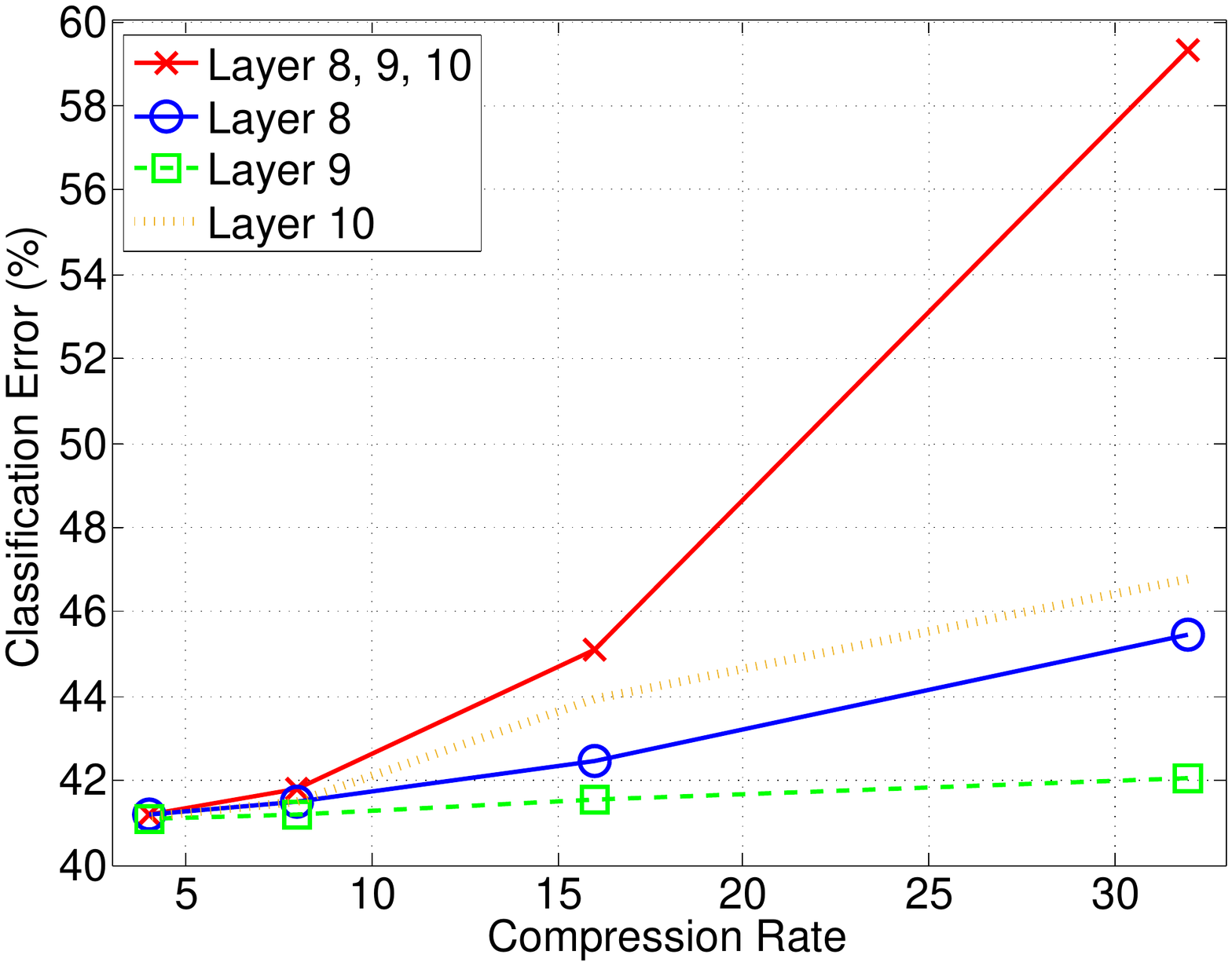}}  
\subfigure[PQ]{
  \includegraphics[width= 1.7in,trim = 10mm 62mm 10mm 75mm]{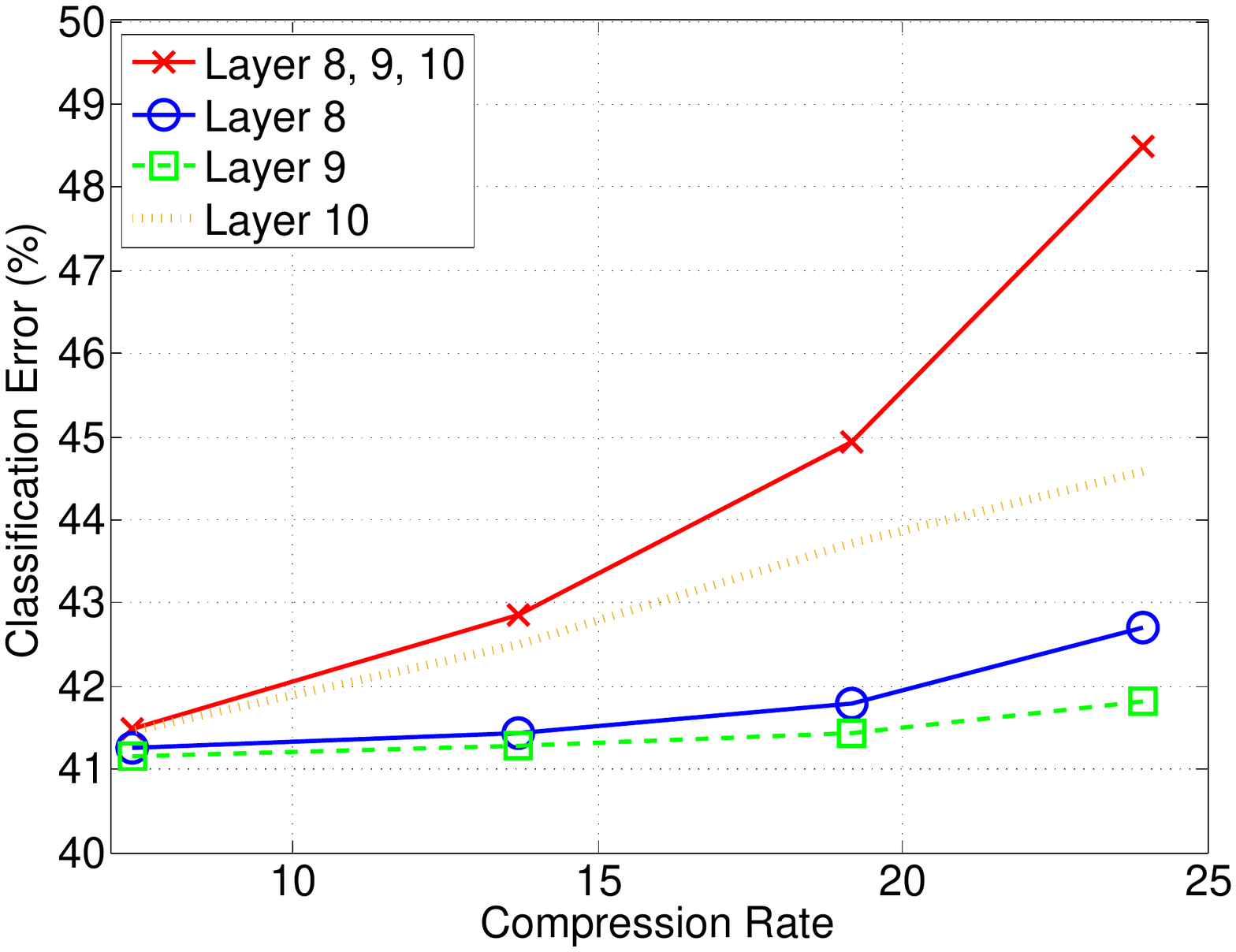}}
  \caption{Comparison of different compression methods when certain layers are compressed and others are fixed to original.}\label{layer}
\end{figure*}

\begin{figure*}[t]
\centering
  \includegraphics[width= 5in,clip=true,trim = 0mm 30mm 10mm 30mm]{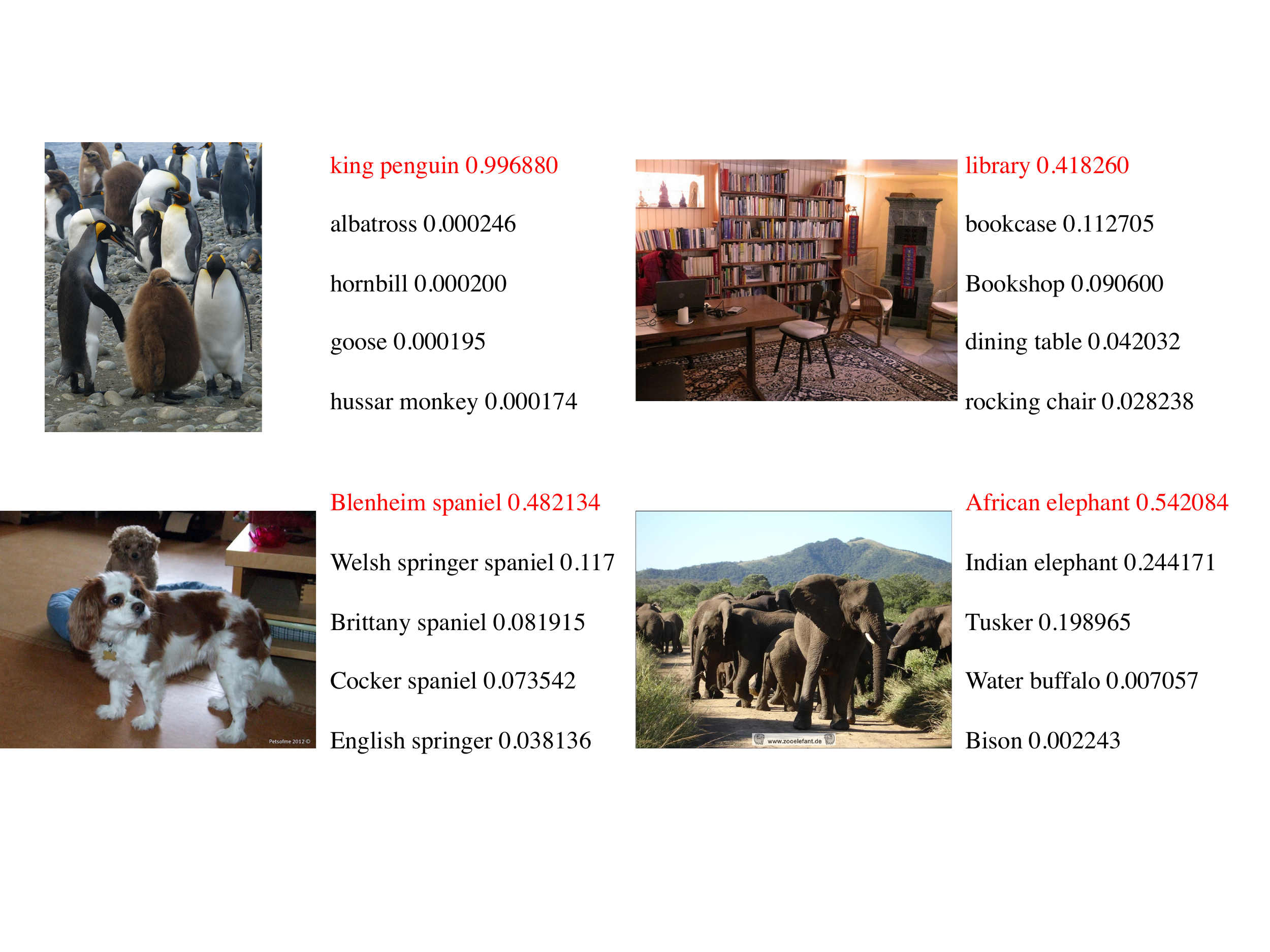} \vspace{-3mm}
  \caption{Some sample predictions results for the PQ compressed CNN.}\label{sample}
\end{figure*}

We also conducted additional analysis on the classification error rate for compressing each single layer while fixing other layers as uncompressed. The results are reported in Figure \ref{layer} (for accuracy@1 only). We found that compressing the eighth and ninth hidden layers did not usually lead to significant decrease of performance, but that compressing the tenth and final classification layer led to a much larger decrease of accuracy. Compressing all three layers together usually led to larger error, especially when the compression rate was high. Finally, some sample predictions results are shown in Figure \ref{sample}.

\section{Application to Image Retrieval}

This section presents an application of the compressed CNN to image retrieval, in order to verify the generalization ability of the compressed networks. In real industry applications, many situations would not allow uploading of a photo to the server, or in which uploading large numbers of photos to the server would be unaffordable. In fact, due to bandwidth limits, we were only able to upload some processed data (e.g., features or hashes of the image) to the server. Given the compressed CNN above, we were able to process the images using compressed CNN on the cellphone side, and to perform retrieval on the database side by uploading only the processed feature.

We performed experiments on the Holidays dataset \citep{Jegou08}. This is a widely used standard benchmark dataset for image retrieval that contains 1491 images for 500 different instances; each instance contains 2-3 images. The number of query images is fixed at 500; the rest are used to populate the database.  We used mean average precision (mAP) to evaluate different methods. We used each compressed CNN model to generate the 2048-dimensional activation features from the last hidden layer, and used them as features for image retrieval. We used cosine distance to measure the similarities among image features.

According to the results in Table \ref{holiday}, the trend of different methods is similar to the classification results; i.e., PQ consistently worked better than the other methods. One surprising finding was that $k$means with 2 centers (1 bit) gave high results--even higher than the original feature (we also verified the variance as 0.1\%). This is a special case, however, which probably came about because the application was near-duplicate image retrieval and because quantizing the values to binary is more robust to small image transformations. However, because our goal was to best reconstruct the original weight matrix, this improvement from this binary $k$means case indeed showed that it is not a very accurate approximation. We are not reporting the results for RQ here, because its performance was so poor.

In conclusion, we found that vector quantized CNNs can be safely applied to problems other than image classification.

\begin{table}[t]
\begin{center}\small
\begin{tabular}{c|c|ccccc}
\hline
 method   	&  ~~compression rate~~         &        ~~mAP~~                \\ \hline
   Original  &     1         &               66.43                            \\ \hline
      Binary  &     32         &                   64.58                        \\ \hline
         SVD  &     32         &               55.49                            \\
   SVD   &     16        &                    63.08                       \\
   SVD   &     8         &                      65.34                     \\
   SVD   &     2         &                      66.23                     \\ \hline  
   KM (2 centers)  &     32         &                    67.61                       \\
   KM (4 centers)  &     16        &                     64.58                      \\
   KM (16 centers)  &     8         &                    66.06                       \\
   KM (32 centers)  &     1         &                   66.58                         \\ \hline
    PQ (2 centers)  &     23.9         &                    64.94                       \\
   PQ (4 centers)  &      13.7     &                         66.37                  \\
   PQ (16 centers)  &     7.4        &                          66.16                 \\
   PQ (32 centers)  &     1         &                        66.40                   \\ \hline
\end{tabular}
\vspace{2mm}
\caption{Results (mAP) on the Holidays dataset for image retrieval. The features were generated using different compressed CNN models.
\label{holiday}}
\end{center}
\end{table}

\section{Discussion}

We have addressed the storage problem of applying vector quantization to compress deep convolutional neural networks in embedded systems. Our work systematically studied how to compress the $10^8$ parameters of a deep convolutional neural network, in order to save storage of the models. Unlike previous approaches that considered using matrix factorization methods, we proposed to study a series of vector quantization methods for compressing the parameters. Somewhat surprisingly, we found that by simply performing a scalar quantization to the parameter values using $k$means, we were able to obtain 8-16 compression rate of the parameters without sacrificing top-five accuracy in more than 0.5\% of the compressions. In addition, by using structured quantization methods, we were able to further compress the parameters up to 24 times while keeping the loss of top-five accuracy within 1\%.

By compressing the parameters more than 20 times, we addressed the problem of applying state-of-the art CNNs in embedded devices. Given a state-of-the-art model with about 200MB of parameters, we were able to reduce them to less than 10MB, which enabled us to easily deploy such models. Another interesting implication of this paper is that our empirical results confirmed the finding in \cite{NIPS2013_5025} that the useful parameters in a CNN are about 5\% (we were able to compress them about 20 times).  

In the future, it will be interesting to explore hardware efficient operations for the compressed models on embedded devices to speed up computation. It will also be interesting to apply fine-tuning to the compressed layers to improve performance. Whereas this paper mainly focused on compressing dense connected layers, it will be interesting to investigate if we can apply the same vector quantization methods to compress convolutional layers.

{
\bibliographystyle{iclr2015}
\bibliography{egbib}
}

\end{document}